\documentclass[11pt,a4paper]{article}
\usepackage[hyperref]{acl2018}
\usepackage{times}
\usepackage{latexsym}

\usepackage{url}
\usepackage{color}
\usepackage{graphicx}
\usepackage{amsmath}
\usepackage{amssymb}
\usepackage{tabularx}
\usepackage{multirow}
\usepackage{mathtools}

\aclfinalcopy 

\makeatletter
\newcommand{\printfnsymbol}[1]{%
  \textsuperscript{\@fnsymbol{#1}}%
}
\makeatother

\title{$N$-ary Relation Extraction using Graph State LSTM}

\author{Linfeng Song$^{1}$\thanks{~~Equal contribution}, Yue Zhang$^{3}$\printfnsymbol{1}, Zhiguo Wang$^2$ \and Daniel Gildea$^1$ \\
  $^1$Department of Computer Science, University of Rochester, Rochester, NY 14627 \\
  $^2$IBM T.J. Watson Research Center, Yorktown Heights, NY 10598 \\
  $^3$School of Engineering, Westlake University, China}

\date{}

\begin{document}
\maketitle
\begin{abstract}
Cross-sentence $n$-ary relation extraction detects relations among $n$ entities across multiple sentences.
Typical methods formulate an input as a \textit{document graph}, integrating various intra-sentential and inter-sentential dependencies.
The current state-of-the-art method splits the input graph into two DAGs, adopting a DAG-structured LSTM for each.
Though being able to model rich linguistic knowledge by leveraging graph edges, important information can be lost in the splitting procedure.
We propose a graph-state LSTM model, which uses a parallel state to model each word, recurrently enriching state values via message passing.
Compared with DAG LSTMs, our graph LSTM keeps the original graph structure, and speeds up computation by allowing more parallelization.
On a standard benchmark, our model shows the best result in the literature.
\end{abstract}

\section{Introduction}
\label{sec:intro}

As a central task in natural language processing, relation extraction has been investigated on news, web text and biomedical domains.
It has been shown to be useful for detecting explicit facts, such as cause-effect \cite{hendrickx2009semeval}, and predicting the effectiveness of a medicine on a cancer caused by mutation of a certain gene in the biomedical domain \cite{quirk-poon:2017:EACLlong,TACL1028}.
While most existing work extracts relations within a sentence \cite{zelenko2003kernel,palmer2005proposition,zhao-grishman:2005:ACL,jiang-zhai:2007:main,plank-moschitti:2013:ACL2013,li-ji:2014:P14-1,gormley-yu-dredze:2015:EMNLP,miwa-bansal:2016:P16-1,zhang-zhang-fu:2017:EMNLP2017}, 
the task of cross-sentence relation extraction has received increasing attention \cite{gerber-chai:2010:ACL,yoshikawa2011coreference}. 
Recently, \newcite{TACL1028} extend cross-sentence relation extraction by further detecting relations among several entity mentions ($n$-ary relation).
Table \ref{tab:task_example} shows an example, which conveys the fact that cancers caused by the \emph{858E} mutation on \emph{EGFR} gene can respond to the \emph{gefitinib} medicine.
The three entity mentions form a ternary relation yet appear in distinct sentences.

\begin{table}
\centering
\begin{tabularx}{0.48\textwidth}{|X|}
\hline
The deletion mutation on exon-19 of \textbf{EGFR} gene was present in 16 patients, while the \textbf{858E} point mutation on exon-21 was noted in 10. \\
All patients were treated with \textbf{gefitinib} and showed a partial response. \\
\hline
\end{tabularx}
\caption{An example showing that tumors with \textit{L858E} mutation in \textit{EGFR} gene respond to \textit{gefitinib} treatment.}
\label{tab:task_example}
\end{table}

\begin{figure*}
\centering
\includegraphics[width=0.9\textwidth]{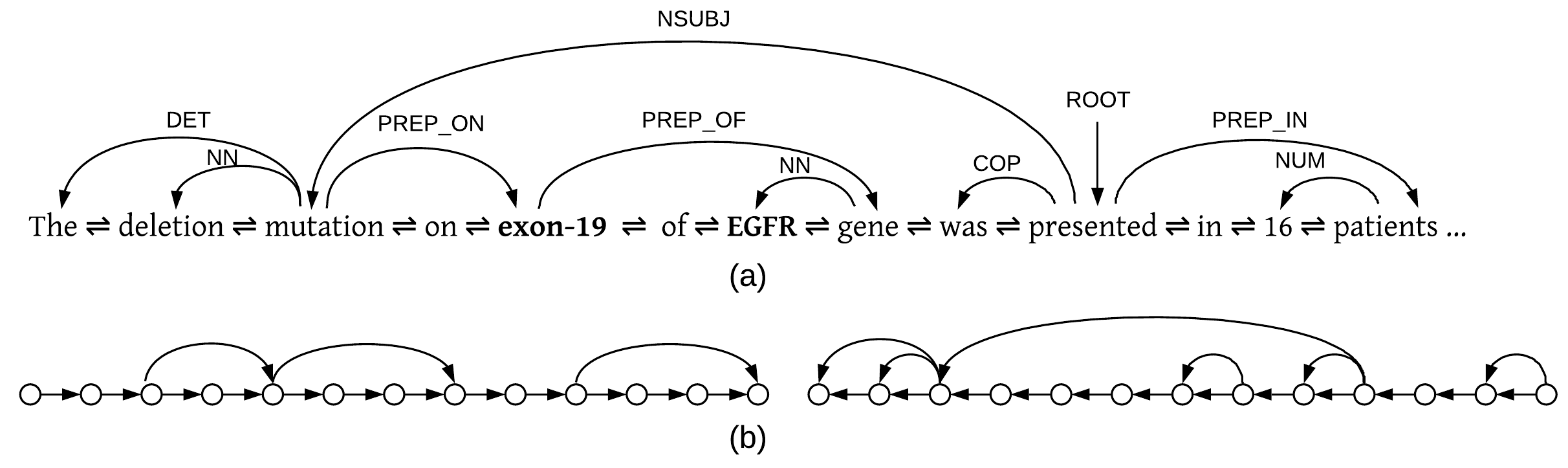}
\caption{(a) A fraction of the dependency graph of the example in Table \ref{tab:task_example}. For simplicity, we omit edges of discourse relations. (b) Results after splitting the graph into two DAGs.}
\label{fig:example_bidir}
\end{figure*}

\newcite{TACL1028} proposed a graph-structured LSTM for $n$-ary relation extraction.
As shown in Figure \ref{fig:example_bidir} (a), graphs are constructed from input sentences with dependency edges, links between adjacent words, and inter-sentence relations, so that syntactic and discourse information can be used for relation extraction.
To calculate a hidden state encoding for each word, \newcite{TACL1028} first split the input graph into two directed acyclic graphs (DAGs) by separating left-to-right edges from right-to-left edges (Figure \ref{fig:example_bidir} (b)).
Then, two separate gated recurrent neural networks, which extend tree LSTM \cite{tai-socher-manning:2015:ACL-IJCNLP}, were adopted for each single-directional DAG, respectively.
Finally, for each word, the hidden states of both directions are concatenated as the final state.
The bi-directional DAG LSTM model showed superior performance 
over several strong baselines, such as tree-structured LSTM \cite{miwa-bansal:2016:P16-1}, 
on a biomedical-domain benchmark.

However, the bidirectional DAG LSTM model suffers from several limitations.
First, important information can be lost when converting a graph into two separate DAGs.
For the example in Figure \ref{fig:example_bidir}, the conversion breaks the inner structure of ``exon-19 of EGFR gene'', 
where the relation between ``exon-19'' and ``EGFR'' via the dependency path ``exon-19 $\xrightarrow{\text{PREP\_OF}}$ gene $\xrightarrow{\text{NN}}$ EGFR'' is lost from the original subgraph.
Second, using LSTMs on both DAGs, information of only ancestors and descendants can be incorporated for each word.
Sibling information, which may also be important, is not included.

A potential solution to the problems above is to model a graph as a whole, learning its representation without breaking it into two DAGs. 
Due to the existence of cycles, naive extension of tree LSTMs cannot serve this goal.
Recently, graph convolutional networks (GCN) \citep{kipf2017semi,marcheggiani-titov:2017:EMNLP2017,bastings-EtAl:2017:EMNLP2017} and graph recurrent networks (GRN) \cite{song-EtAl:acl2018,zhang-EtAl:acl2018} have been proposed for representing graph structures for NLP tasks.
Such methods encode a given graph by hierarchically learning representations of neighboring nodes in the graphs via their connecting edges.
While GCNs use CNN for information exchange, GRNs take gated recurrent steps to this end. 
For fair comparison with DAG LSTMs, we build a graph LSTM by extending \newcite{song-EtAl:acl2018}, which strictly follow the configurations of \newcite{TACL1028} such as the source of features and hyper parameter settings.
In particular, the full input graph is modeled as a single state, with words in the graph being its sub states. 
State transitions are performed on the graph recurrently, allowing word-level states to exchange information through dependency and discourse edges. 
At each recurrent step, each word advances its current state by receiving information from the current states of its adjacent words. 
Thus with increasing numbers of recurrent steps each word receives information from a larger context.
Figure \ref{fig:transition} shows the recurrent transition steps where each node works simultaneously within each transition step.

Compared with bidirectional DAG LSTM, our method has several advantages.
First, it keeps the original graph structure, and therefore no information is lost.
Second, sibling information can be easily incorporated by passing information up and then down from a parent.
Third, information exchange allows more parallelization, and thus can be very efficient in computation.

Results show that our model outperforms a bidirectional DAG LSTM baseline by 5.9\% in accuracy,
overtaking the state-of-the-art system of \newcite{TACL1028} 
by 1.2\%.
Our code is available at \url{https://github.com/freesunshine0316/nary-grn}.

Our contributions are summarized as follows. 
\begin{itemize}
\item We empirically compared graph LSTM with DAG LSTM for $n$-ary relation extraction tasks, showing that the former is better by more effective use of structural information;
\item To our knowledge, we are the first to investigate a graph recurrent network for modeling dependency and discourse relations.
\end{itemize}

\section{Task Definition}
\label{sec:task}

Formally, the input for cross-sentence $n$-ary relation extraction can be represented as a pair $(\mathcal{E}, \mathcal{T})$, where $\mathcal{E} = (\epsilon_1, \dots, \epsilon_N)$ is the set of entity mentions, and $\mathcal{T} = [S_1;\dots;S_M]$ is a text consisting of multiple sentences.
Each entity mention $\epsilon_i$ belongs to one sentence in $\mathcal{T}$.
There is a predefined relation set $\mathcal{R} = (r_1, \dots, r_L, \textit{None})$, where \textit{None} represents that no relation holds for the entities.
This task can be formulated as a binary classification problem of determining whether $\epsilon_1, \dots, \epsilon_N$ together form a relation \cite{TACL1028}, or a multi-class classification problem of detecting which relation holds for the entity mentions.  
Take Table \ref{tab:task_example} as an example. 
The binary classification task is to determine whether \emph{gefitinib} would have an effect on this type of cancer, given a cancer patient with \emph{858E} mutation on gene \emph{EGFR}.
The multi-class classification task is to detect the exact drug effect: response, resistance, sensitivity, etc.

\section{Baseline: Bi-directional DAG LSTM}
\label{sec:baseline}

\newcite{TACL1028} formulate the task as a graph-structured problem in order to adopt rich dependency and discourse features.
In particular, Stanford parser \cite{manning-EtAl:2014:P14-5} is used to assign syntactic structure to input sentences, and heads of two consecutive sentences are connected to represent discourse information, resulting in a graph structure.
For each input graph $G=(V,E)$, the nodes $V$ are words within input sentences, and each edge $e \in E$  connects two words that either have a relation or are adjacent to each other.
Each edge is denoted as a triple $(i,j,l)$, where $i$ and $j$ are the indices of the source and target words, respectively, and the edge label $l$ indicates either a dependency or discourse relation (such as ``nsubj'') or a relative position (such as ``next\_tok'' or ``prev\_tok'').
Throughout this paper, we use $E_{in}(j)$ and $E_{out}(j)$ to denote the sets of incoming and outgoing edges for word $j$.

For a bi-directional DAG LSTM baseline, we follow \newcite{TACL1028}, splitting each input graph into two separate DAGs by separating left-to-right edges from right-to-left edges (Figure \ref{fig:example_bidir}).
Each DAG is encoded by using a DAG LSTM (Section \ref{sec:baseline_gated}), which takes both source words and edge labels as inputs (Section \ref{sec:baseline_input}).
Finally, the hidden states of entity mentions from both LSTMs are taken as inputs to a logistic regression classifier to make a prediction:
\begin{equation}
\hat{y} = \textrm{softmax}(W_0 [h_{\epsilon_1}; \dots; h_{\epsilon_N}] + b_0) \textrm{,}
\end{equation}
where $h_{\epsilon_j}$ is the hidden state of entity $\epsilon_j$.
$W_0$ and $b_0$ are parameters.

\subsection{Input Representation}
\label{sec:baseline_input}

Both nodes and edge labels are useful for modeling a syntactic graph.
As the input to our DAG LSTM, we first calculate the representation for each edge $(i,j,l)$ by:
\begin{equation} \label{eq:baseline_edge}
x_{i,j}^l = W_1 \Big( [e_l; e_i] \Big) + b_1 \textrm{,}
\end{equation}
where $W_1$ and $b_1$ are model parameters, $e_i$ is the embedding of the source word indexed by $i$, and $e_l$ is the embedding of the edge label $l$.

\subsection{State transition}
\label{sec:baseline_gated}

The baseline LSTM model learns DAG representations sequentially, following word orders.
Taking the edge representations (such as $x_{i,j}^l$) as input, gated state transition operations are executed on both the forward and backward DAGs.
For each word $j$, the representations of its incoming edges $E_{in}(j)$ are summed up as one vector:
\begin{equation} \label{eq:baseline_input}
x_j^{in} = \sum_{(i,j,l)\in E_{in}(j)} x_{i,j}^l \\
\end{equation}
Similarly, for each word $j$, the states of all incoming nodes are summed to a single vector before being passed to the gated operations:
\begin{equation}
h_j^{in} = \sum_{(i,j,l)\in E_{in}(j)} h_{i} \\
\end{equation}
Finally, the gated state transition operation for the hidden state $h_j$ of the $j$-th word can be defined as:
\vspace{0.0em}
\begin{equation} \label{eq:baseline_gated}
\begin{split}
i_j &= \sigma(W_i x_j^{in} + U_i h_j^{in} + b_i) \\
o_j &= \sigma(W_o x_j^{in} + U_o h_j^{in} + b_o) \\
f_{i,j} &= \sigma(W_f x_{i,j}^l + U_f h_i + b_f) \\
u_j &= \sigma(W_u x_j^{in} + U_u h_j^{in} + b_u) \\
c_j &= i_j \odot u_j + \sum_{(i,j,l)\in E_{in}(j)} f_{i,j} \odot c_i \\
h_j &= o_j \odot \tanh (c_j) \textrm{,} \\
\end{split}
\end{equation}
where $i_j$, $o_j$ and $f_{i,j}$ are a set of input, output and forget gates, respectively, and $W_x$, $U_x$ and $b_x$ ($x \in \{i,o,f,u\}$) are model parameters.

\subsection{Comparison with \newcite{TACL1028}}
\label{sec:baseline_comp}

Our baseline is computationally similar to \newcite{TACL1028}, but different on how to utilize edge labels in the gated network.
In particular, \newcite{TACL1028} make model parameters specific to edge labels. 
They consider two model variations, namely \emph{Full Parametrization (FULL)} and \emph{Edge-Type Embedding (EMBED)}.
\emph{FULL} assigns distinct $U$s (in Equation \ref{eq:baseline_gated}) to different edge types, so that each edge label is associated with a 2D weight matrix to be tuned in training.
On the other hand, \emph{EMBED} assigns each edge label to an embedding vector, but complicates the gated operations by changing the $U$s to be 3D tensors.\footnote{For more information please refer Section 3.3 of \newcite{TACL1028}.}

In contrast, we take edge labels as part of the input to the gated network.
In general, the edge labels are first represented as embeddings, before being concatenated with the node representation vectors (Equation \ref{eq:baseline_edge}).
We choose this setting for both the baseline and our graph state LSTM model in Section \ref{sec:model}, since it requires fewer parameters compared with \emph{FULL} and \emph{EMBED}, thus being less exposed to overfitting on small-scaled data.

\section{Graph State LSTM}
\label{sec:model}

Our input graph formulation strictly follows Section \ref{sec:baseline}. 
In particular, our model adopts the same methods for calculating input representation (as in Section \ref{sec:baseline_input}) and performing classification as the baseline model.
However, different from the baseline bidirectional DAG LSTM model, we leverage a graph-structured LSTM to directly model the input graph, without splitting it into two DAGs.

\begin{figure}
\centering
\includegraphics[width=0.45\textwidth]{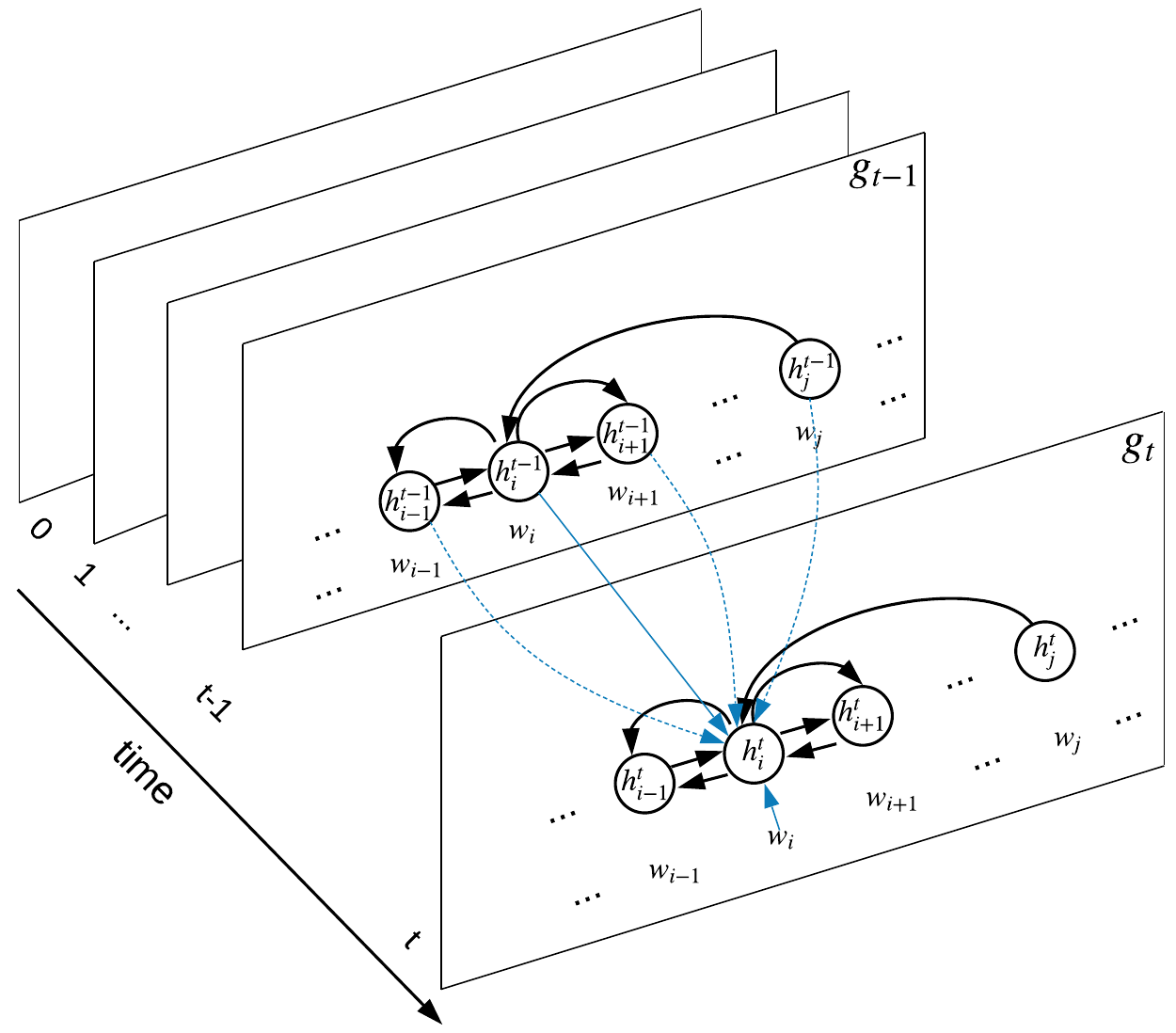}
\caption{Graph state transitions via message passing, where each $w_i$ is a word.}
\label{fig:transition}
\end{figure}

Figure \ref{fig:transition} shows an overview of our model. 
Formally, given an input graph $G=(V, E)$, we define a state vector $h^j$ for each word $v_j \in V$. 
The state of the graph consists of all word states, and thus can be represented as:
\begin{equation}
g = \{h^j\}|_{v_j \in V}
\end{equation}
In order to capture non-local information, our model performs information exchange between words through a recurrent state transition process, resulting in a sequence of graph states $g_0, g_1, \dots, g_t$, where $g_t = \{h_t^j\}|_{v_j \in V}$.
The initial graph state $g_0$ consists of a set of initial word states $h_0^j=h_0$, where $h_0$ is a zero vector.

\subsection{State transition}

Following the approches of \newcite{song-EtAl:acl2018} and \newcite{zhang-EtAl:acl2018}, a recurrent neural network is utilized to model the state transition process. 
In particular, the transition from $g_{t-1}$ to $g_t$ consists of hidden state transition for each word, as shown in Figure \ref{fig:transition}. 
At each step $t$, we allow information exchange between a word and all words that are directly connected to the word. 
To avoid gradient diminishing or bursting, gated LSTM cells are adopted, where a cell $c_t^j$ is taken to record memory for $h_t^j$. 
We use an input gate $i_t^j$, an output gate $o_t^j$ and a forget gate $f_t^j$ to control information flow from the inputs and to $h_t^j$.

The inputs to a word $v_j$, include representations of edges that are connected to $v_j$, where $v_j$ can be either the source or the target of the edge.
Similar to Section \ref{sec:baseline_input}, we define each edge as a triple $(i,j,l)$, where $i$ and $j$ are indices of the source and target words, respectively, and $l$ is the edge label.
$x_{i,j}^l$ is the representation of edge $(i,j,l)$.
The inputs for $v_j$ are distinguished by incoming and outgoing directions, where:
\begin{equation}
\begin{split}
x_j^{i} &= \sum_{(i,j,l)\in E_{in}(j)} x_{i,j}^l \\
x_j^{o} &= \sum_{(j,k,l)\in E_{out}(j)} x_{j,k}^l \\
\end{split}
\end{equation}
Here $E_{in}(j)$ and $E_{out}(j)$ denote the sets of incoming and outgoing edges of $v_j$, respectively.

In addition to edge inputs, a cell also takes the hidden states of its incoming and outgoing words during a state transition. 
In particular, the states of all incoming words and outgoing words are summed up, respectively:
\begin{equation}
\begin{split}
h_j^{i} &= \sum_{(i,j,l)\in E_{in}(j)} h_{t-1}^{i} \\
h_j^{o} &= \sum_{(j,k,l)\in E_{out}(j)} h_{t-1}^{k} \textrm{,} \\
\end{split}
\end{equation} 
Based on the above definitions of $x_j^{i}$, $x_j^{o}$, $h_j^{i}$ and $h_j^{o}$, the recurrent state transition from $g_{t-1}$ to $g_t$, as represented by $h_t^j$, is defined as:
\begin{equation*}
\begin{split}
i_t^j &= \sigma(W_i x_j^{i} + \hat{W_i} x_j^{o} + U_i h_j^{i} + \hat{U_i} h_j^{o} + b_i) \\
o_t^j &= \sigma(W_o x_j^{i} + \hat{W_o} x_j^{o} + U_o h_j^{i} + \hat{U_o} h_j^{o} + b_o) \\
f_t^j &= \sigma(W_f x_j^{i} + \hat{W_f} x_j^{o} + U_f h_j^{i} + \hat{U_f} h_j^{o} + b_f) \\
u_t^j &= \sigma(W_u x_j^{i} + \hat{W_u} x_j^{o} + U_u h_j^{i} + \hat{U_u} h_j^{o} + b_u) \\
c_t^j &= f_t^j \odot c_{t-1}^j + i_t^j \odot u_t^j \\
h_t^j &= o_t^j \odot \tanh (c_t^j) \textrm{,} \\
\end{split}
\end{equation*}
where $i_t^j$, $o_t^j$ and $f_t^j$ are the input, output and forget gates, respectively. $W_x$, $\hat{W}_x$, $U_x$, $\hat{U}_x$, $b_x$ ($x \in \{i, o, f, u\}$) are model parameters.


\subparagraph{Graph State LSTM vs bidirectional DAG LSTM}
A contrast between the baseline DAG LSTM and our graph LSTM can be made from the perspective of information flow. 
For the baseline, information flow follows the natural word order in the input sentence, with the two DAG components propagating information from left to right and from right to left, respectively. 
In contrast, information flow in our graph state LSTM is relatively more concentrated at individual words, with each word exchanging information with all its graph neighbors simultaneously at each sate transition. 
As a result, wholistic contextual information can be leveraged for extracting features for each word, as compared to separated handling of bi-directional information flow in DAG LSTM. 
In addition, arbitrary structures, including arbitrary cyclic graphs, can be handled.

From an initial state with isolated words, information of each word propagates to its graph neighbors after each step. 
Information exchange between non-neighboring words can be achieved through multiple transition steps.
We experiment with different transition step numbers to study the effectiveness of global encoding. 
Unlike the baseline DAG LSTM encoder, our model allows parallelization in node-state updates, and thus can be highly efficient using a GPU.

\section{Training}

We train our models with a cross-entropy loss over a set of gold standard data:
\begin{equation}
l = -\log p(y_i|X_i;\theta)\textrm{,}
\end{equation}
where $X_i$ is an input graph, $y_i$ is the gold class label of $X_i$, and $\theta$ is the model parameters.
Adam \cite{kingma2014adam} with a learning rate of 0.001 is used as the optimizer, and the model that yields the best devset performance is selected to evaluate on the test set.
Dropout with rate 0.3 is used during training.
Both training and evaluation are conducted using a Tesla K20X GPU.

\section{Experiments}

We conduct experiments for the binary relation detection task and the multi-class relation extraction task discussed in Section \ref{sec:task}.

\subsection{Data}
\label{sec:data}

\begin{table} 
\centering
\begin{tabular} {l|c|c|c}
\hline
Data & Avg. Tok. & Avg. Sent. & Cross \\
\hline
\textsc{Ternary} & 73.9 & 2.0 & 70.1\% \\ 
\textsc{Binary}  & 61.0 & 1.8 & 55.2\% \\
\hline
\end{tabular}
\caption{Dataset statistics. \emph{Avg.} \emph{Tok.} and \emph{Avg.} \emph{Sent.} are the average number of tokens and sentences, respectively. \emph{Cross} is the percentage of instances that contain multiple sentences.}
\label{tab:stat}
\end{table}

We use the dataset of \newcite{TACL1028}, which is a biomedical-domain dataset focusing on drug-gene-mutation ternary relations,\footnote{The dataset is available at \\ \url{http://hanover.azurewebsites.net}.} extracted from PubMed.
It contains 6987 ternary instances about drug-gene-mutation relations, and 6087 binary instances about drug-mutation sub-relations.
Table \ref{tab:stat} shows statistics of the dataset. 
Most instances of ternary data contain multiple sentences, and the average number of sentences is around 2.
There are five classification labels: ``resistance or non-response'', ``sensitivity'', ``response'', ``resistance'' and ``None''.
We follow \newcite{TACL1028} and binarize multi-class labels by grouping all relation classes as ``Yes'' and treat ``None'' as ``No''.

\subsection{Settings}

Following \newcite{TACL1028}, five-fold cross-validation is used for evaluating the models,\footnote{The released data has been separated into 5 portions, and we follow the exact split.}
and the final test accuracy is calculated by averaging the test accuracies over all five folds.
For each fold, we randomly separate 200 instances from the training set for development.
The batch size is set as 8 for all experiments. 
Word embeddings are initialized with the 100-dimensional GloVe \cite{pennington2014glove} vectors, pretrained on 6 billion words from Wikipedia and web text.
The edge label embeddings are 3-dimensional and randomly initialized.
Pretrained word embeddings are not updated during training.
The dimension of hidden vectors in LSTM units is set to 150.

\subsection{Development Experiments}

\begin{figure}
\centering
\includegraphics[width=0.45\textwidth]{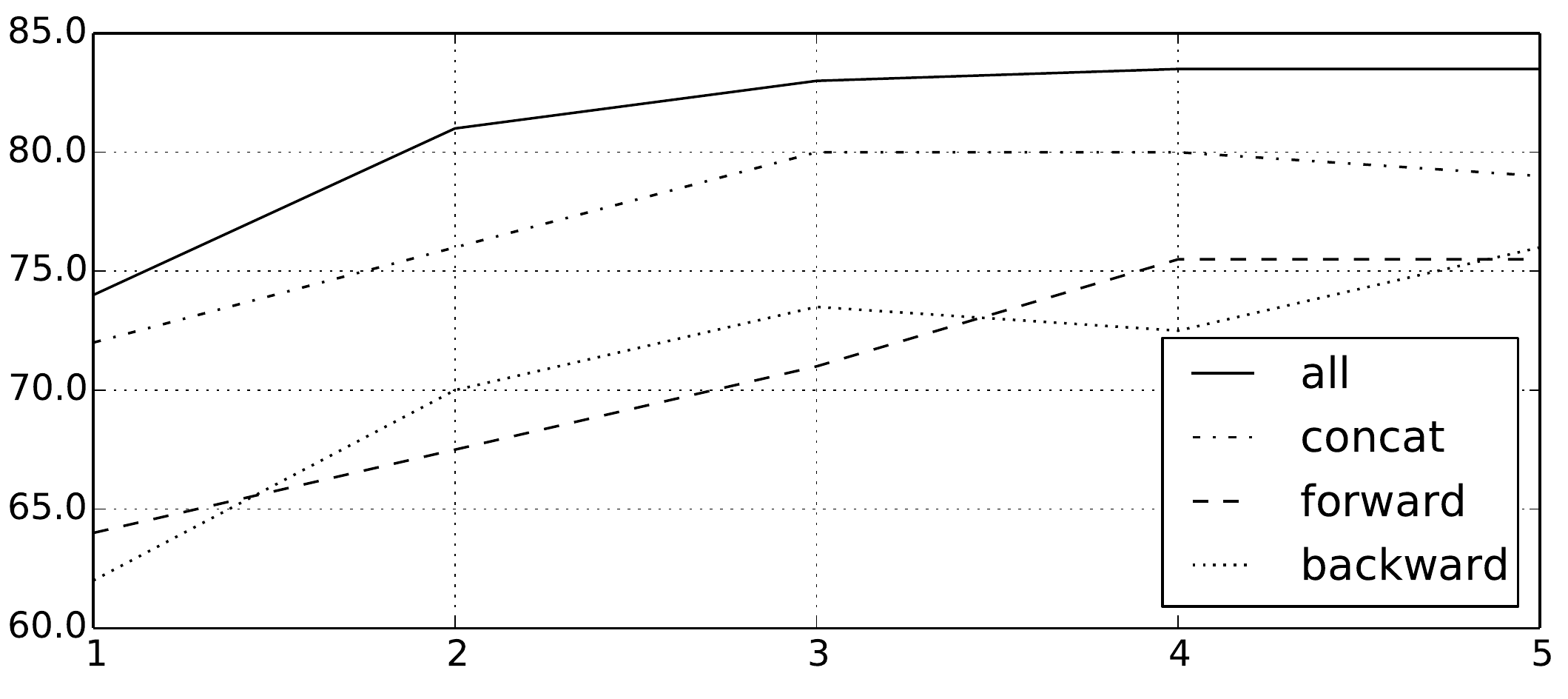}
\caption{Dev accuracies against transition steps for the graph state LSTM model.}
\label{fig:dev}
\end{figure}

We first analyze our model on the drug-gene-mutation ternary relation dataset, taking the first among 5-fold cross validation settings for our data setting.
Figure \ref{fig:dev} shows the devset accuracies of different state transition numbers, where \emph{forward} and \emph{backward} execute our graph state model only on the forward or backward DAG, respectively.
\emph{Concat} concatenates the hidden states of \emph{forward} and \emph{backward}.
\emph{All} executes our graph state model on original graphs.

The performance of \emph{forward} and \emph{backward} lag behind \emph{concat}, which is consistent with the intuition that both forward and backward relations are useful \cite{TACL1028}.
In addition, \emph{all} gives better accuracies compared with \emph{concat}, demonstrating the advantage of simultaneously considering forward and backward relations during representation learning.
For all the models, more state transition steps result in better accuracies, where larger contexts can be integrated in the representations of graphs.
The performance of \emph{all} starts to converge after 4 and 5 state transitions, so we set the number of state transitions to 5 in the remaining experiments.

\subsection{Final results}
\label{sec:results}

\begin{table}
\centering
\begin{tabular}{lcc}
\hline
Model & Single & Cross  \\
\hline
\newcite{quirk-poon:2017:EACLlong} & 74.7 & 77.7 \\ 
\newcite{TACL1028} - EMBED & 76.5 &  80.6 \\ 
\newcite{TACL1028} - FULL & 77.9 & 80.7 \\ 
~~~~~~~~~~~+ multi-task & -- & 82.0 \\ 
\hline
Bidir DAG LSTM & 75.6 & 77.3 \\
GS GLSTM  & \textbf{80.3*} & \textbf{83.2*}  \\
\hline
\end{tabular}
\caption{Average test accuracies for \textsc{Ternary} drug-gene-mutation interactions. \emph{Single} represents experiments only on instances within single sentences, while \emph{Cross} represents experiments on all instances. *: significant at $p<0.01$}
\label{tab:ternary}
\end{table}

Table \ref{tab:ternary} compares our model with the bidirectional DAG baseline and the state-of-the-art results on this dataset, where \emph{EMBED} and \emph{FULL} have been briefly introduced in Section \ref{sec:baseline_comp}.
\emph{+multi-task} applies joint training of both ternary (drug-gene-mutation) relations and their binary (drug-mutation) sub-relations.
\newcite{quirk-poon:2017:EACLlong} use a statistical method with a logistic regression classifier and features derived from shortest paths between all entity pairs.
\emph{Bidir DAG LSTM} is our bidirectional DAG LSTM baseline, and \emph{GS GLSTM} is our graph state LSTM model.

Using all instances (the \emph{Cross} column in Table \ref{tab:ternary}), our graph state LSTM model shows the highest test accuracy among all methods, which is 5.9\% higher than our baseline.\footnote{$p<0.01$ using t-test. For the remaining of this paper, we use the same measure for statistical significance.}
The accuracy of our baseline is lower than \emph{EMBED} and \emph{FULL} of \newcite{TACL1028}, which is likely due to the differences mentioned in Section \ref{sec:baseline_comp}.
Our final results are better than \newcite{TACL1028}, despite the fact that we do not use multi-task learning.

We also report accuracies only on instances within single sentences (column \emph{Single} in Table \ref{tab:ternary}), which exhibit similar contrasts.
Note that all systems show performance drops when evaluated only on single-sentence relations, which are actually more challenging.
One reason may be that some single sentences cannot provide sufficient context for disambiguation, making it necessary to study cross-sentence context.
Another reason may be overfitting caused by relatively fewer training instances in this setting, as only 30\% instances are within a single sentence. 
One interesting observation is that our baseline shows the least performance drop of 1.7 points, in contrast to up to 4.1 for other neural systems.
This can be a supporting evidence for overfitting, as our baseline has fewer parameters at least than \emph{FULL} and \emph{EMBED}.

\subsection{Analysis}
\label{sec:analysis}

\begin{table}
\centering
\begin{tabular}{lcc}
\hline
Model & Train & Decode \\
\hline
Bidir DAG LSTM & 281s & 27.3s \\
GS GLSTM  & 36.7s & 2.7s \\
\hline
\end{tabular}
\caption{The average times for training one epoch and decoding (seconds) over five folds on drug-gene-mutation \textsc{Ternary} cross sentence setting.}
\label{tab:time}
\end{table}

\subparagraph{Efficiency.} 
Table \ref{tab:time} shows the training and decoding time of both the baseline and our model.
Our model is 8 to 10 times faster than the baseline in training and decoding speeds, respectively.
By revisiting Table \ref{tab:stat}, we can see that the average number of tokens for the ternary-relation data is 74, which means that the baseline model has to execute 74 recurrent transition steps for calculating a hidden state for each input word. 
On the other hand, our model only performs 5 state transitions, and calculations between each pair of nodes for one transition are parallelizable.
This accounts for the better efficiency of our model.

\begin{figure*}
\centering
\includegraphics[width=0.95\textwidth]{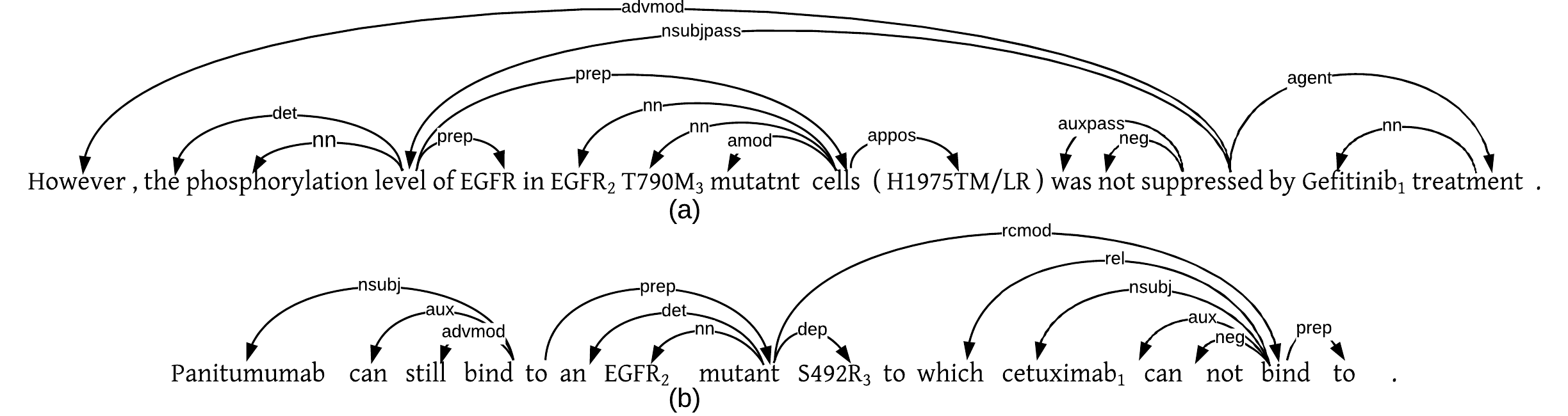}
\caption{Example cases. Words with subindices 1, 2 and 3 represent drugs, genes and mutations, respectively. References for both cases are ``No''. For both cases, \emph{GS GLSTM} makes the correct predictions, while \emph{Bidir DAG LSTM} does incorrectly.}
\label{fig:case_study}
\end{figure*}

\begin{figure}
\centering
\includegraphics[width=0.48\textwidth]{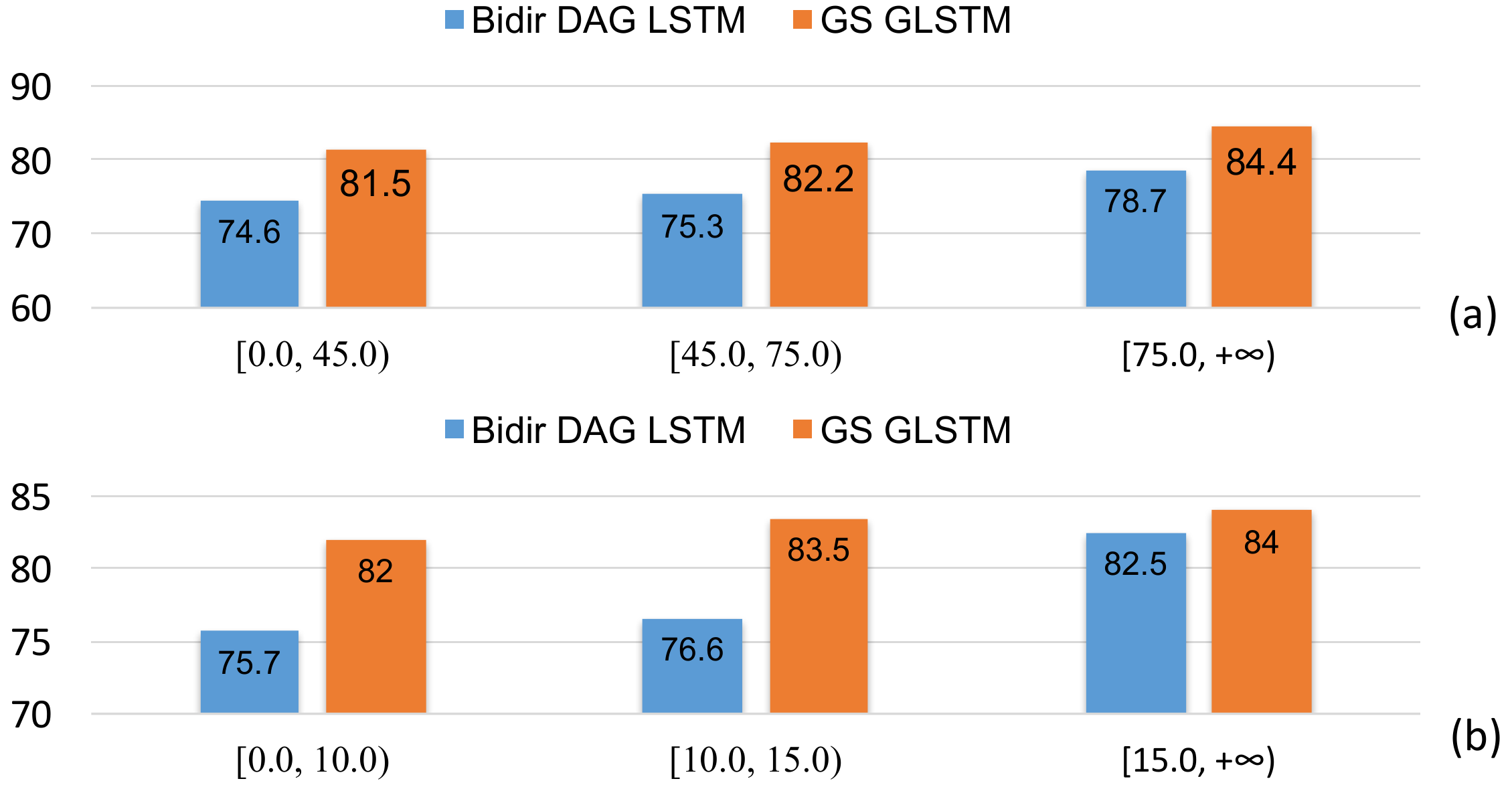}
\caption{Test set performances on (a) different sentence lengths, and (b) different maximal number of neighbors.}
\label{fig:by_length}
\end{figure}

\subparagraph{Accuracy against sentence length}
Figure \ref{fig:by_length} (a) shows the test accuracies on different sentence lengths.
We can see that \emph{GS GLSTM} and \emph{Bidir DAG LSTM} show performance increase along increasing input sentence lengths.
This is likely because longer contexts provide richer information for relation disambiguation.
\emph{GS GLSTM} is consistently better than \emph{Bidir DAG LSTM}, and the gap is larger on shorter instances.
This demonstrates that \emph{GS GLSTM} is more effective in utilizing a smaller context for disambiguation.

\subparagraph{Accuracy against the maximal number of neighbors}
Figure \ref{fig:by_length} (b) shows the test accuracies against the maximum number of neighbors.
Intuitively, it is easier to model graphs containing nodes with more neighbors, because these nodes can serve as a ``supernode'' that allow more efficient information exchange. 
The performances of both \emph{GS GLSTM} and \emph{Bidir DAG LSTM} increase with increasing maximal number of neighbors, which coincide with this intuition.
In addition, \emph{GS GLSTM} shows more advantage than \emph{Bidir DAG LSTM} under the inputs having lower maximal number of neighbors, which further demonstrates the superiority of \emph{GS GLSTM} over \emph{Bidir DAG LSTM} in utilizing context information.

\subparagraph{Case study}
Figure \ref{fig:case_study} visualizes the merits of \emph{GS GLSTM} over \emph{Bidir DAG LSTM} using two examples. 
\emph{GS GLSTM} makes the correct predictions for both cases, while \emph{Bidir DAG LSTM} fails to.

The first case generally mentions that \emph{Gefitinib} does not have an effect on \emph{T790M} mutation on \emph{EGFR} gene.
Note that both ``However'' and ``was not'' serve as indicators; thus incorporating them into the contextual vectors of these entity mentions is important for making a correct prediction.
However, both indicators are leaves of the dependency tree, making it impossible for \emph{Bidir DAG LSTM} to incorporate them into the contextual vectors of entity mentions up the tree through dependency edges.\footnote{As shown in Figure \ref{fig:example_bidir}, a directional DAG LSTM propagates information according to the edge directions.}
On the other hand, it is easier for \emph{GS GLSTM}.
For instance, ``was not'' can be incorporated into ``Gefitinib'' through ``suppressed $\xrightarrow{\text{agent}}$ treatment $\xrightarrow{\text{nn}}$ Gefitinib''.

\begin{table}
\centering
\begin{tabular}{lcccc}
\hline
Model & Single & Cross \\
\hline
\newcite{quirk-poon:2017:EACLlong} & 73.9 & 75.2 \\
\newcite{miwa-bansal:2016:P16-1} & 75.9 & 75.9 \\ 
\newcite{TACL1028} - EMBED & 74.3 &  76.5 \\ 
\newcite{TACL1028} - FULL & 75.6 & 76.7 \\ 
~~~~~~~~~~~+ multi-task & -- & 78.5 \\ 
\hline
Bidir DAG LSTM & 76.9 & 76.4 \\
GS GLSTM  & \textbf{83.5*} & \textbf{83.6*} \\
\hline
\end{tabular}
\caption{Average test accuracies in five-fold cross-validation for \textsc{Binary} drug-mutation interactions.}
\label{tab:binary}
\end{table}

The second case is to detect the relation among ``cetuximab'' (drug), ``EGFR'' (gene) and ``S492R'' (mutation), which does not exist.
However, the context introduces further ambiguity by mentioning another drug ``Panitumumab'', which does have a  relation with ``EGFR'' and ``S492R''.
Being sibling nodes in the dependency tree, ``can not'' is an indicator for the relation of ``cetuximab''.
\emph{GS GLSTM} is correct, because ``can not'' can be easily included into the contextual vector of ``cetuximab'' in two steps via ``bind $\xrightarrow{\text{nsubj}}$cetuximab''.

\subsection{Results on Binary Sub-relations}

Following previous work, we also evaluate our model on drug-mutation binary relations.
Table \ref{tab:binary} shows the results, where \newcite{miwa-bansal:2016:P16-1} is a state-of-the-art model using sequential and tree-structured LSTMs to jointly capture linear and dependency contexts for relation extraction.
Other models have been introduced in Section \ref{sec:results}.

Similar to the ternary relation extraction experiments, \emph{GS GLSTM} outperforms all the other systems with a large margin, which shows that the message passing graph LSTM is better at encoding rich linguistic knowledge within the input graphs.
Binary relations being easier, both \emph{GS GLSTM} and \emph{Bidir DAG LSTM} show increased or similar performances compared with the ternary relation experiments. 
On this set, our bidirectional DAG LSTM model is comparable to \emph{FULL} using all instances (``Cross'') and slightly better than \emph{FULL} using only single-sentence instances (``Single'').

\subsection{Fine-grained Classification}

Our dataset contains five classes as mentioned in Section \ref{sec:data}.
However, previous work only investigates binary relation detection.
Here we also study the multi-class classification task, which can be more informative for applications.

Table \ref{tab:multi_class} shows accuracies on multi-class relation extraction, which makes the task more ambiguous compared with binary relation extraction.
The results show similar comparisons with the binary relation extraction results.
However, the performance gaps between \emph{GS GLSTM} and \emph{Bidir DAG LSTM} dramatically increase, showing the superiority of \emph{GS GLSTM} over \emph{Bidir DAG LSTM} in utilizing context information.

\section{Related Work}

\subparagraph{$N$-ary relation extraction}
$N$-ary relation extractions can be traced back to MUC-7 \cite{chinchor1998overview}, which focuses on entity-attribution relations.
It has also been studied in biomedical domain \cite{mcdonald-EtAl:2005:ACL}, but only the instances within a single sentence are considered.
Previous work on cross-sentence relation extraction relies on either explicit co-reference annotation \cite{gerber-chai:2010:ACL,yoshikawa2011coreference}, or the assumption that the whole document refers to a single coherent event \cite{wick-culotta-mccallum:2006:EMNLP,swampillai-stevenson:2011:RANLP}.
Both simplify the problem and reduce the need for learning better contextual representation of entity mentions.
A notable exception is \newcite{quirk-poon:2017:EACLlong}, who adopt distant supervision and integrated contextual evidence of diverse types without relying on these assumptions.
However, they only study binary relations.
We follow \newcite{TACL1028} by studying ternary cross-sentence relations.

\begin{table}
\centering
\begin{tabular}{lcc}
\hline
Model & \textsc{Ternary} & \textsc{Binary} \\
\hline
Bidir DAG LSTM & 51.7 & 50.7 \\
GS GLSTM & \textbf{71.1*} & \textbf{71.7*} \\
\hline
\end{tabular}
\caption{Average test accuracies for multi-class relation extraction with all instances (``Cross'').}
\label{tab:multi_class}
\end{table}

\textbf{Graph encoder}
\newcite{liang2016semantic} build a graph LSTM model for semantic object parsing, which aims to segment objects within an image into more fine-grained, semantically meaningful parts.
The nodes of an input graph come from image superpixels, and the edges are created by connecting spatially neighboring nodes.
Their model is similar as \newcite{TACL1028} by calculating node states sequentially: for each input graph, a start node and a node sequence are chosen, which determines the order of recurrent state updates.
In contrast, our graph LSTM do not need ordering of graph nodes, and is highly parallelizable.

Graph convolutional networks (GCNs) and very recently graph recurrent networks (GRNs) have been used to model graph structures in NLP tasks, such as semantic role labeling \cite{marcheggiani-titov:2017:EMNLP2017}, machine translation \cite{bastings-EtAl:2017:EMNLP2017}, text generation \cite{song-EtAl:acl2018}, text representation \cite{zhang-EtAl:acl2018} and semantic parsing \cite{Exploitingkun2018,xu2018graph2seq}.
In particular, \newcite{zhang-EtAl:acl2018} use GRN to represent raw sentences by building a graph structure of neighboring words and a sentence-level node, showing that the encoder outperforms BiLSTMs and Transformer \cite{NIPS2017_7181} on classification and sequence labeling tasks; 
\newcite{song-EtAl:acl2018} build a GRN for encoding AMR graphs, showing that the representation is superior compared to BiLSTM on serialized AMR. 
Our work is in line with their work in the investigation of GRN on NLP. 
To our knowledge, we are the first to use GRN for representing dependency and discourse structures.
Under the same recurrent framework, we show that modeling the original graphs with one GRN model is more useful than two DAG LSTMs for our relation extraction task.
We choose GRN as our main method because it gives a more fair comparison with DAG LSTM.
We leave it to future work to compare GCN and GRN for our task.


\section{Conclusion}

We explored a graph-state LSTM model for cross-sentence $n$-ary relation extraction, which uses a recurrent state transition process to incrementally refine a neural graph state representation capturing graph structure contexts.
Compared with a bidirectional DAG LSTM baseline, our model has several advantages.
First, it does not change the input graph structure, so that no information can be lost.
For example, it can easily incorporate sibling information when calculating the contextual vector of a node.
Second, it is better parallelizable.
Experiments show significant improvements over the previously reported numbers, including that of the bidirectional graph LSTM model.

For future work, we consider adding coreference information as an entity mention can have coreferences, which help on information collection.
Another possible direction is including word sense information.
Confusing caused by word senses can be a severe problem.
Not only content words, but also propositions can introduce word sense problem \cite{gong2018embedding}.

\subparagraph{Acknowledge}
We thank the anonymized reviewers for their insightful comments, and the Center for Integrated Research Computing (CIRC) of University of Rochester for making special reservations for computation resources.

\bibliography{acl2018}

\begin{thebibliography}{}
\expandafter\ifx\csname natexlab\endcsname\relax\def\natexlab#1{#1}\fi

\bibitem[{Bastings et~al.(2017)Bastings, Titov, Aziz, Marcheggiani, and
  Simaan}]{bastings-EtAl:2017:EMNLP2017}
Joost Bastings, Ivan Titov, Wilker Aziz, Diego Marcheggiani, and Khalil Simaan.
  2017.
\newblock Graph convolutional encoders for syntax-aware neural machine
  translation.
\newblock In {\em Conference on Empirical Methods in Natural Language
  Processing (EMNLP-17)\/}.

\bibitem[{Chinchor(1998)}]{chinchor1998overview}
Nancy~A Chinchor. 1998.
\newblock Overview of muc-7/met-2.
\newblock Technical report, SCIENCE APPLICATIONS INTERNATIONAL CORP SAN DIEGO
  CA.

\bibitem[{Gerber and Chai(2010)}]{gerber-chai:2010:ACL}
Matthew Gerber and Joyce Chai. 2010.
\newblock Beyond nombank: A study of implicit arguments for nominal predicates.
\newblock In {\em Proceedings of the 48th Annual Meeting of the Association for
  Computational Linguistics (ACL-10)\/}.

\bibitem[{Gong et~al.(2018)Gong, Bhat, and Viswanath}]{gong2018embedding}
Hongyu Gong, Suma Bhat, and Pramod Viswanath. 2018.
\newblock Embedding syntax and semantics of prepositions via tensor
  decomposition.
\newblock In {\em Proceedings of the 2018 Meeting of the North American chapter
  of the Association for Computational Linguistics (NAACL-18)\/}.

\bibitem[{Gormley et~al.(2015)Gormley, Yu, and
  Dredze}]{gormley-yu-dredze:2015:EMNLP}
Matthew~R. Gormley, Mo~Yu, and Mark Dredze. 2015.
\newblock Improved relation extraction with feature-rich compositional
  embedding models.
\newblock In {\em Conference on Empirical Methods in Natural Language
  Processing (EMNLP-15)\/}.

\bibitem[{Hendrickx et~al.(2009)Hendrickx, Kim, Kozareva, Nakov,
  {\'O}~S{\'e}aghdha, Pad{\'o}, Pennacchiotti, Romano, and
  Szpakowicz}]{hendrickx2009semeval}
Iris Hendrickx, Su~Nam Kim, Zornitsa Kozareva, Preslav Nakov, Diarmuid
  {\'O}~S{\'e}aghdha, Sebastian Pad{\'o}, Marco Pennacchiotti, Lorenza Romano,
  and Stan Szpakowicz. 2009.
\newblock Semeval-2010 task 8: Multi-way classification of semantic relations
  between pairs of nominals.
\newblock In {\em Proceedings of the Workshop on Semantic Evaluations: Recent
  Achievements and Future Directions\/}.

\bibitem[{Jiang and Zhai(2007)}]{jiang-zhai:2007:main}
Jing Jiang and ChengXiang Zhai. 2007.
\newblock A systematic exploration of the feature space for relation
  extraction.
\newblock In {\em Proceedings of the 2015 Meeting of the North American chapter
  of the Association for Computational Linguistics (NAACL-15)\/}.

\bibitem[{Kingma and Ba(2014)}]{kingma2014adam}
Diederik Kingma and Jimmy Ba. 2014.
\newblock Adam: A method for stochastic optimization.
\newblock {\em arXiv preprint arXiv:1412.6980\/} .

\bibitem[{Kipf and Welling(2017)}]{kipf2017semi}
Thomas~N. Kipf and Max Welling. 2017.
\newblock Semi-supervised classification with graph convolutional networks.
\newblock In {\em International Conference on Learning Representations
  (ICLR)\/}.

\bibitem[{Li and Ji(2014)}]{li-ji:2014:P14-1}
Qi~Li and Heng Ji. 2014.
\newblock Incremental joint extraction of entity mentions and relations.
\newblock In {\em Proceedings of the 52nd Annual Meeting of the Association for
  Computational Linguistics (ACL-14)\/}.

\bibitem[{Liang et~al.(2016)Liang, Shen, Feng, Lin, and
  Yan}]{liang2016semantic}
Xiaodan Liang, Xiaohui Shen, Jiashi Feng, Liang Lin, and Shuicheng Yan. 2016.
\newblock Semantic object parsing with graph {LSTM}.
\newblock In {\em European Conference on Computer Vision\/}.

\bibitem[{Manning et~al.(2014)Manning, Surdeanu, Bauer, Finkel, Bethard, and
  McClosky}]{manning-EtAl:2014:P14-5}
Christopher~D. Manning, Mihai Surdeanu, John Bauer, Jenny Finkel, Steven~J.
  Bethard, and David McClosky. 2014.
\newblock The {Stanford} {CoreNLP} natural language processing toolkit.

\bibitem[{Marcheggiani and Titov(2017)}]{marcheggiani-titov:2017:EMNLP2017}
Diego Marcheggiani and Ivan Titov. 2017.
\newblock Encoding sentences with graph convolutional networks for semantic
  role labeling.
\newblock In {\em Conference on Empirical Methods in Natural Language
  Processing (EMNLP-17)\/}.

\bibitem[{McDonald et~al.(2005)McDonald, Pereira, Kulick, Winters, Jin, and
  White}]{mcdonald-EtAl:2005:ACL}
Ryan McDonald, Fernando Pereira, Seth Kulick, Scott Winters, Yang Jin, and Pete
  White. 2005.
\newblock Simple algorithms for complex relation extraction with applications
  to biomedical {IE}.
\newblock In {\em Proceedings of the 43rd Annual Meeting of the Association for
  Computational Linguistics (ACL'05)\/}.

\bibitem[{Miwa and Bansal(2016)}]{miwa-bansal:2016:P16-1}
Makoto Miwa and Mohit Bansal. 2016.
\newblock End-to-end relation extraction using {LSTM}s on sequences and tree
  structures.
\newblock In {\em Proceedings of the 54th Annual Meeting of the Association for
  Computational Linguistics (ACL-16)\/}.

\bibitem[{Palmer et~al.(2005)Palmer, Gildea, and
  Kingsbury}]{palmer2005proposition}
Martha Palmer, Daniel Gildea, and Paul Kingsbury. 2005.
\newblock The proposition bank: An annotated corpus of semantic roles.
\newblock {\em Computational linguistics\/} 31(1):71--106.

\bibitem[{Peng et~al.(2017)Peng, Poon, Quirk, Toutanova, and Yih}]{TACL1028}
Nanyun Peng, Hoifung Poon, Chris Quirk, Kristina Toutanova, and Wen-tau Yih.
  2017.
\newblock Cross-sentence n-ary relation extraction with graph {LSTM}s.
\newblock {\em Transactions of the Association for Computational Linguistics\/}
  5:101--115.

\bibitem[{Pennington et~al.(2014)Pennington, Socher, and
  Manning}]{pennington2014glove}
Jeffrey Pennington, Richard Socher, and Christopher~D. Manning. 2014.
\newblock {GloVe}: Global vectors for word representation.
\newblock In {\em Conference on Empirical Methods in Natural Language
  Processing (EMNLP-14)\/}.

\bibitem[{Plank and Moschitti(2013)}]{plank-moschitti:2013:ACL2013}
Barbara Plank and Alessandro Moschitti. 2013.
\newblock Embedding semantic similarity in tree kernels for domain adaptation
  of relation extraction.
\newblock In {\em Proceedings of the 51st Annual Meeting of the Association for
  Computational Linguistics (ACL-13)\/}.

\bibitem[{Quirk and Poon(2017)}]{quirk-poon:2017:EACLlong}
Chris Quirk and Hoifung Poon. 2017.
\newblock Distant supervision for relation extraction beyond the sentence
  boundary.
\newblock In {\em Proceedings of the 15th Conference of the European Chapter of
  the ACL (EACL-17)\/}.

\bibitem[{Song et~al.(2018)Song, Zhang, Wang, and Gildea}]{song-EtAl:acl2018}
Linfeng Song, Yue Zhang, Zhiguo Wang, and Daniel Gildea. 2018.
\newblock A graph-to-sequence model for amr-to-text generation.
\newblock In {\em Proceedings of the 56th Annual Meeting of the Association for
  Computational Linguistics (ACL-18)\/}.

\bibitem[{Swampillai and Stevenson(2011)}]{swampillai-stevenson:2011:RANLP}
Kumutha Swampillai and Mark Stevenson. 2011.
\newblock Extracting relations within and across sentences.
\newblock In {\em Proceedings of the International Conference Recent Advances
  in Natural Language Processing 2011\/}.

\bibitem[{Tai et~al.(2015)Tai, Socher, and
  Manning}]{tai-socher-manning:2015:ACL-IJCNLP}
Kai~Sheng Tai, Richard Socher, and Christopher~D. Manning. 2015.
\newblock Improved semantic representations from tree-structured long
  short-term memory networks.
\newblock In {\em Proceedings of the 53rd Annual Meeting of the Association for
  Computational Linguistics (ACL-15)\/}.

\bibitem[{Vaswani et~al.(2017)Vaswani, Shazeer, Parmar, Uszkoreit, Jones,
  Gomez, Kaiser, and Polosukhin}]{NIPS2017_7181}
Ashish Vaswani, Noam Shazeer, Niki Parmar, Jakob Uszkoreit, Llion Jones,
  Aidan~N Gomez, \L~ukasz Kaiser, and Illia Polosukhin. 2017.
\newblock Attention is all you need.
\newblock In I.~Guyon, U.~V. Luxburg, S.~Bengio, H.~Wallach, R.~Fergus,
  S.~Vishwanathan, and R.~Garnett, editors, {\em Advances in Neural Information
  Processing Systems 30\/}, pages 5998--6008.

\bibitem[{Wick et~al.(2006)Wick, Culotta, and
  McCallum}]{wick-culotta-mccallum:2006:EMNLP}
Michael Wick, Aron Culotta, and Andrew McCallum. 2006.
\newblock Learning field compatibilities to extract database records from
  unstructured text.
\newblock In {\em Conference on Empirical Methods in Natural Language
  Processing (EMNLP-06)\/}.

\bibitem[{Xu et~al.(2018{\natexlab{a}})Xu, Wu, Wang, and
  Sheinin}]{xu2018graph2seq}
Kun Xu, Lingfei Wu, Zhiguo Wang, and Vadim Sheinin. 2018{\natexlab{a}}.
\newblock Graph2seq: Graph to sequence learning with attention-based neural
  networks.
\newblock {\em arXiv preprint arXiv:1804.00823\/} .

\bibitem[{Xu et~al.(2018{\natexlab{b}})Xu, Wu, Wang, Yu, Chen, and
  Sheinin}]{Exploitingkun2018}
Kun Xu, Lingfei Wu, Zhiguo Wang, Mo~Yu, Liwei Chen, and Vadim Sheinin.
  2018{\natexlab{b}}.
\newblock Exploiting rich syntactic information for semantic parsing with
  graph-to-sequence model.
\newblock In {\em Conference on Empirical Methods in Natural Language
  Processing (EMNLP-18)\/}.

\bibitem[{Yoshikawa et~al.(2011)Yoshikawa, Riedel, Hirao, Asahara, and
  Matsumoto}]{yoshikawa2011coreference}
Katsumasa Yoshikawa, Sebastian Riedel, Tsutomu Hirao, Masayuki Asahara, and
  Yuji Matsumoto. 2011.
\newblock Coreference based event-argument relation extraction on biomedical
  text.
\newblock {\em Journal of Biomedical Semantics\/} 2(5):S6.

\bibitem[{Zelenko et~al.(2003)Zelenko, Aone, and
  Richardella}]{zelenko2003kernel}
Dmitry Zelenko, Chinatsu Aone, and Anthony Richardella. 2003.
\newblock Kernel methods for relation extraction.
\newblock {\em Journal of machine learning research\/} 3(Feb):1083--1106.

\bibitem[{Zhang et~al.(2017)Zhang, Zhang, and
  Fu}]{zhang-zhang-fu:2017:EMNLP2017}
Meishan Zhang, Yue Zhang, and Guohong Fu. 2017.
\newblock End-to-end neural relation extraction with global optimization.
\newblock In {\em Conference on Empirical Methods in Natural Language
  Processing (EMNLP-17)\/}.

\bibitem[{Zhang et~al.(2018)Zhang, Liu, and Song}]{zhang-EtAl:acl2018}
Yue Zhang, Qi~Liu, and Linfeng Song. 2018.
\newblock Sentence-state lstm for text representation.
\newblock In {\em Proceedings of the 56th Annual Meeting of the Association for
  Computational Linguistics (ACL-18)\/}.

\bibitem[{Zhao and Grishman(2005)}]{zhao-grishman:2005:ACL}
Shubin Zhao and Ralph Grishman. 2005.
\newblock Extracting relations with integrated information using kernel
  methods.
\newblock In {\em Proceedings of the 43rd Annual Meeting of the Association for
  Computational Linguistics (ACL'05)\/}.

\end{thebibliography}
\bibliographystyle{acl_natbib}

\end{document}